# REASON AGAINST THE MACHINE
## FUTURE DIRECTIONS FOR MASS ONLINE DELIBERATION

**Ruth Shortall**
TU Delft

**Anatol Itten**
TU Delft

**Michiel van der Meer**
Leiden University

**Pradeep K. Murukannaiah**
TU Delft

**Catholijn M. Jonker**
TU Delft, Leiden University

ABSTRACT

Designers of online deliberative platforms aim to counter the degrading quality of online debates. Support technologies such as machine learning and natural language processing open avenues for widening the circle of people involved in deliberation, moving from small groups to "crowd" scale. Numerous design features of large-scale online discussion systems allow larger numbers of people to discuss shared problems, enhance critical thinking, and formulate solutions. We review the transdisciplinary literature on the design of digital mass deliberation platforms and examine the commonly featured design aspects (e.g., argumentation support, automated facilitation, and gamification) that attempt to facilitate scaling up. We find that the literature is largely focused on developing technical fixes for scaling up deliberation, but may neglect the more nuanced requirements of high quality deliberation. Current design research is carried out with a small, atypical segment of the world's population, and much research is still needed on how to facilitate and accommodate different genders or cultures in deliberation, how to deal with the implications of pre-existing social inequalities, how to build motivation and self-efficacy in certain groups, and how to deal with differences in cognitive abilities and cultural or linguistic differences. Few studies bridge disciplines between deliberative theory, design and engineering. As a result, scaling up deliberation will likely advance in separate systemic siloes. We make design and process recommendations to correct this course and suggest avenues for future research.

*Keywords* digital deliberation · design · automated facilitation · argumentation tools · gamification

## 1 Introduction

While social media and other platforms allow mass participation and sharing of political opinion, they limit exposure to opposing views (Kim et al., 2019), have been shown to create echo chambers or filter bubbles (Bozdag and Van Den Hoven, 2015), and are not effective in reducing the spread of conspiracy content (Faddoul et al., 2020). Furthermore, the policies and design characteristics of popular online platforms are frequently blamed for proliferating discriminatory and abusive behaviour (Wulczyn et al. 2017), due to the way they shape our interactions (Levy and Barocas, 2017). Platforms like Facebook or Twitter generally respond to offensive material reactively, censoring it (some argue, to the detriment of free expression) only after complaints are received, which is too late to undo psychological harm to the recipients (Ullmann and Tomalin, 2020). These social media platforms are clearly unsuited to enabling respectful and reasoned discussions around urgent, important, controversial and complex systemic challenges like climate change adaptation or migration (Gürkan et al., 2010), which are outside the realms of routine experience.

Deliberation offers a decision-making approach for such complex policy problems (Wironen et al., 2019; Dryzek et al. 2019). Deliberative exercises, such as mini-publics, are largely held with randomly selected small groups in both online and offline settings, and offer sufficient time to reflect on arguments, pose questions, or collaborate on solutions. In order to promote effective deliberation online, researchers have proposed that discussion platforms be designed in accordance with deliberative ideals, to increase equity and inclusiveness (e.g. Zhang, 2010) and eliminate discriminatory effects of class, race, and gender inequalities (Gutmann et al., 2004). Not only should they promote a greater degree of equality or civility between participants (Gastil and Black, 2007), they should also redistribute power among ordinary citizens (Curato et al., 2019). Deliberative online platforms ideally strive to promote respectful and thoughtful discussion. Their potential to reduce polarization, build civic capacity and produce higher quality opinions (Strandberg and Grönlund, 2012) is much discussed in the literature. Design features of platforms that promote deliberation are shown to be evaluated more favourably by citizens (Christensen, 2021).

Deliberative discussions, in the most general sense, can be held between any group of people. If based on a randomly selected mini-public model, Goodin (2000) argues, they can accurately reflect the views of a larger group had the process been carried out at that scale. They can provide a certain guarantee of 'representativeness' with inclusion of minority voices (Lafont, 2019; Curato et al., 2019). The number of participants in deliberative exercises is typically limited to several hundred at most to maintain organizational feasibility. However, Fishkin (2020) argues that those who do not participate in deliberations are likely to be 'disengaged and inattentive', and are not encouraged to think about the complexities of policy issues posing difficult trade-offs. Simply put, the fewer the people deliberating, the easier it will be to uphold high deliberative standards; but this limits high civic engagement. Conversely, the more people deliberating, the more difficult it will be to safeguard high deliberative standards; but this makes it possible to reach a high civic engagement.

However, obtaining high citizen engagement *and* maintaining high deliberative standards is particularly appealing now that there is a growing demand for citizen participation in urgent and important policy issues such as climate change or the energy transition (Schleifer and Diep, 2019). And even more so, because valuable and vital public ideas are inadequately reflected in current small-group deliberative practices (Yang et al., 2021). Indeed, the distribution of people's ideas is shown to have a long tail, thus requiring the participation of masses of people to ensure the diversity of ideas is adequately captured (Klein, 2012). There is further empirical evidence that participants from the wider public process the objective information presented in deliberative mini-publics quite differently than the members of this mini-public itself (Suiter et al., 2020). Promoting high-quality mass deliberation online is arguably essential to enhance critical thinking and reflection, to build greater understanding of diverse perspectives and policy issues among participants, while contributing to widely supported solutions (Gürkan et al., 2010; Verdiesen et al., 2018). However, scaling up online deliberation is challenging.

To begin with, attracting users to deliberate is not easy (Toots, 2019) and there is a myriad of reasons why citizens will not take the time and effort to deliberate in depth. For instance, a lack of participation of citizens in online deliberations is related to how people's own abilities, capacities or one's general perception of online deliberation spaces (Jacquet, 2017). Online deliberation spaces may have high barriers to entry for the public (Epstein and Leshed, 2016) and are "typically one-off experiments that occur within the confines of a single issue over a short period of time" (Leighninger, 2012). Logically, the initial focus of designers of crowd-deliberation platforms at this stage is certainly to reduce thresholds to participate, and increase activity and engagement to deliberation. Various challenges to scaling up deliberation will need to be overcome, such as how to manage and facilitate discussion between large numbers (Hartz-Karp and Sullivan, 2014) of users and allow them to make sense of each other's arguments, how to synthesise the large amount of data that arises (Ercan et al., 2019), how to deal with potential loss of social cues (Rhee and Kim, 2009), how to attract and engage an adequate number of users and thus ensure that platform design upholds democratic ideals such as inclusion, equality, and reflection or representativeness (Alnemr, 2020).

Online deliberative spaces remain in the experimental phase and come in all shapes and sizes with no clear consensus about what is the best model. Numerous design solutions to challenges of scaling up have been proposed in the literature to date. Previous reviews e.g. (Jonsson and Åström, 2014) reviewed research on online deliberation and how the concept of deliberation is interpreted by researchers; (Friess and Eilders, 2015) analysed the state of research on online deliberation according to a framework of institutional input, quality of the communication process the expected results of deliberation. While these reviews provide a broad overview of research themes in the online deliberation field, and while design features are discussed to some extent (e.g. mode of communication, anonymity, moderation) neither provide an up-to-date comprehensive systematic overview of existing platform design features. Nor does, to our knowledge, any comprehensive up-to-date review of literature on *mass* deliberative platforms and their design characteristics exist. In this review, we therefore explicitly focus on design features of deliberative online platforms that are developed for large groups.

Indeed, Friess and Eilders (2015) point out that the design component of online deliberation needs to be better understood, especially how design influences the outcomes of deliberation (i.e. effects on the individual as well as quality of the deliberation). Previous work has investigated the extent to which online deliberation fulfils various deliberative ideals, such as civility, respect and heterogeneity, argumentative quality, reflexivity or inclusiveness (Rossini et al., 2021). More recently, Gastil and Broghammer (2021) call for more research on the impact of digital deliberative platforms on institutional legitimacy, which arises from perceptions of procedural justice and trust. We argue that these perceptions are directly influenced by the design features and design process used to create the platform as well as how the chosen design characteristics promote inclusion and equality during deliberation.

In this paper, we primarily aim to analyse the literature with regard to the focus of designers and their progress in addressing the challenges to scaling up deliberation. As well as this, with regard to the potential impact of digital platforms on the legitimacy of deliberation, our secondary goal is to shed light on the issues of inclusion and equality in the design of these platforms. We believe this will provide valuable insights for the future design and uptake of online deliberative platforms.

This paper is structured as follows: Section 2 describes the challenges involved in scaling up deliberation. Section 3 presents the method used for the systematic review. In Section 4, we categorize our findings on design aspects according to how they address challenges of scaling up deliberation. Section 5 critically discusses these findings, and offers some points of reflection. Finally, we lay out our conclusions in Section 6.

## 2 Challenges to scaling up deliberation online

Some deliberative scholars conclude that the values of deliberation and large scale participation, and hence the combination thereof - mass deliberation - are inextricably in conflict (Cohen, 2009; Lafont, 2019). Lafont for example notes that publicity could seriously undermine the quality of deliberation as it could attract attention of populist forces or trolls without a genuine interest into the deliberation itself. Moreover, publicity might lead participants to appear tougher and be less cooperative than they really are (Ross and Ward, 1995).

Furthermore, since deliberative processes have been designed to encourage slow and careful reasoning, at a large scale they may become too slow to deal with urgent policy problems, such as the ecological crises that we are facing presently, which demand immediate, urgent action (Wironen et al. 2019). In the current set-up it is understood that the greater the number of people that participate in a deliberation, the less likely it is for everyone to have equal time to explain their views, ask questions and receive answers, weigh new considerations and the like (Lafont, 2019).

Individually, humans are shown to be poor reasoners, although their argumentative capabilities improve when they are encouraged to communicate (Chambers, 2018). Ercan et al. (2019) argue that in a society with abundant information and communication, an important design consideration for online deliberation is to encourage listening and reflection, followed by decision-making. However, there is a trade-off between user accessibility and an understandable, well-structured discussion that encourages reflection. Most online discussions happen on easy-to-use conversation-based platforms like forums, even though their ability to promote fair and transparent discussion is debatable (Klein, 2012; Black et al., 2011; Fishkin et al., 2018). For example, the structure of comments sections in news media websites is shown to affect the deliberative quality of discussions (Peacock et al., 2019). Another example are the political discourses under a Facebook comment that are of significant low deliberative quality (Fournier-Tombs and Di Marzo Serugendo, 2020). Posts organised temporally, rather than topically are more difficult to navigate and connect to each other and content tends to be repeated. In mass deliberation, without appropriate support, it is impossible for humans to understand and synthesise the large amounts of information that result let alone reflect on them. Furthermore, disorganised opinions, which may not always be based on facts, make it difficult to identify and understand the arguments of others. These could be structured by means of argument structuring or reasoning tools (Verdiesen et al., 2018). Visualizing discussions and mapping out arguments helps participants to clarify their thinking and better connect information (Popa et al., 2020; Gürkan and Iandoli, 2009). Such tools or platforms may require user training or supervision, but they also counter sponsored content and promote fair and rational assessment of alternatives (Iandoli et al., 2018). Argumentation tools are therefore a promising avenue for allowing quicker navigation of arguments for participants and encouraging reasoned reflection.

Furthermore, having an independent moderator can vastly improve the quality of any discussion, since they can enforce social norms and deliberative ideals. For example, Ito (2018) finds a positive effect on discussion quality due to the 'social presence' of a facilitator in the online platform, along with facilitator support functions. However, facilitation and moderation is also a challenge when scaling up deliberation. For example, Klein (2012) estimates requiring a human moderator for every 20 users, which theoretically severely limits the ability to scale up to large crowds. Human-annotations and facilitating larger scale online deliberations can be challenging, is prohibitively expensive and resource-intensive (Wulczyn et al. 2017). With mass deliberation, moderator workload becomes too high so some aspects of facilitation may need to be automated to deal with tasks like opinion summarisation or consensus building (Lee et al., 2020). Fishkin et al. (2018) also note a problem with scaling up to be finding recruiting and training neutral (human) moderators. As well as time and location constraints, human moderators also suffer from human bias. Beck et al. (2018), for instance, show how human moderator's beliefs and values may introduce bias into discussions. Support technologies stemming from Artificial Intelligence (AI), such as Natural Language processing (NLP), have been used to aid the moderation and related argument mapping procedures.

Reduced face-to-face interaction may lead to a loss of social cues and empathy between participants. A lack of ability to hear, see, or share physical space between participants increases the required effort to communicate and may result in lower mutual understanding (Iandoli et al., 2014). A loss of social cues may also result in a loss of accountability or respectfulness (Sarmento and Mendonça, 2014).

A further challenge relates to the risk that the perceived legitimacy of deliberative platforms may suffer since they not uphold certain democratic ideals, such as inclusion, representativeness, equity or equality, which they claim to promote (Alnemr, 2020). Lack of participation in online deliberation has been linked to perceptions of deliberation as perpetuating existing power structures (Neblo et al., 2010). Hence, it remains a challenge to engage the wider population in intensive deliberative processes, and those that will engage can be expected to be unrepresentative of the rest of the population (Fishkin, 2020). In mass deliberation a greater diversity of participants is theoretically possible, however, three main factors of power distortion in online deliberations are said to be gender, social status and knowledge (Monnoyer-Smith & Wojcik, 2012). These factors may not be considered in the design of online deliberation platforms (Epstein et al., 2014). To the contrary, there is a risk that platform designs may actually strengthen pre-existing (unconscious) biases, i.e. prejudice based on gender, ethnicity, race, class or sexual preference. In the results section, we review the strengths and weaknesses of various design features found in the literature that aim to support scaling up deliberation online by addressing some of the challenges above. In this regard, we also discuss the characteristics of case-studies found in the literature, such as the geographical spread and characteristics of test users.

## 3 Method

This studies aims to provide a comprehensive review of the state of the research of online deliberation platform design. A systematic literature review was performed on academic (English language) literature relating to digital deliberation platform design. The review consisted of three main steps: 1) literature search, 2) screening, 3) quantitative and thematic analysis. Several searches for academic literature were performed on Google Scholar and Scopus between August 2020 and October 2021. An overview of the method is given in Figure 1. An initial pool of publications was identified using various combinations of the keywords: 'digital deliberation', 'online deliberation', 'mass', 'design' and by reading the abstract. This pool was further expanded by snowballing on relevant citations, including using 'related literature' links in Google Scholar search. In total 140 publications were found in the literature search. These were screened for relevance by reading in depth by a team of researchers. This reduced the number of publications to 85. Relevance screening criteria for academic articles were as follows:

- Publication was about an online deliberation platform design case study, i.e. it describes the design or proposed design of a digital deliberation tool or platform, or article analyses existing design(s).
- Publication was peer reviewed or from high quality source.

- Access was available via author's institution.

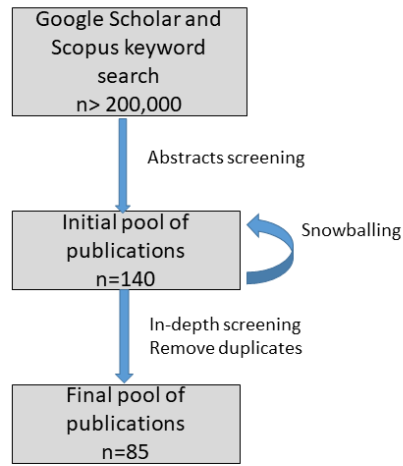

*Figure 1: literature search procedure*

Publications were categorised by first clustering literature into themes. Several major design themes were identified within the pool of publications (Figure 2). The publications were further organised to better reflect how platform designs address the challenges (see Section 2) of mass online deliberation. For case-studies, their characteristics such as number and type of user, study country of origin and research discipline were recorded. Appendix A provides a detailed overview of all case-studies examined in our review. (A list of all reviewed publications is available on request from the first author).

## 4 Results

A total of 85 publications were reviewed. Of these, 62 were about design case studies and the rest analysed existing designs or suggested possible design features, e.g. Wright and Street (2007); Towne and Herbsleb (2012); Ruckenstein and Turunen (2020); Bozdag and Van Den Hoven (2015). The most popular topics were argumentation tools (18 publications), e.g. (Iandoli et al., 2014; Gold et al., 2018) and facilitation (16 publications), e.g. (Wyss and Beste, 2017; Lee et al., 2020). Within the facilitation topic, human facilitation in online deliberation platforms was discussed in 4 publications e.g. (Velikanov, 2017; Epstein and Leshed, 2016).

Gamification was another topic drawing substantial interest with 7 publications, e.g. (Gastil and Broghammer, 2021; Gordon et al., 2016). Other publications covered the topics of media presentation (3 publications) (Brinker et al., 2015; Ramsey and Wilson, 2009; Semaan et al., 2015) or other specific design dimensions. For example, we found work on virtual reality (Gordon and Manosevitch, 2011), visual cues (Manosevitch et al., 2014) or reflection spaces (Ercan et al., 2019) (1 publication each). The remaining publications (34) focused on general platform design or numerous design features in the same paper. Among these publications, other popular themes of discussion included participant anonymity vs. identity e.g. (Gonçalves et al., 2020; Rose and Sæbø, 2010; Rhee and Kim, 2009) and the use of asynchronous vs real-time discussion or text-based vs. video deliberation e.g. (Osborne et al., 2018; Kennedy et al., 2020).

Among the publications dealing with specific case-studies, the USA (24 publications) is dominant in this field of research, followed by Europe (23 publications) and Asia (Japan, Korea, Singapore) (8 publications). The remaining publications were from Russia, Australia, Brazil, Israel and Afghanistan. (Note: some publications involved research in more than one country).

With regard to the spread across disciplines, around half of the 85 publications were published in what may be considered interdisciplinary, multidisciplinary or transdisciplinary journals or proceedings (e.g. *Journal of Information Technology and Politics, Policy and Internet, Government Information Quarterly, New Media and Society, Social Informatics*). The remainder were mainly published in either 'pure' computer either science (e.g. *Applied Intelligence, Journal of Information Science, Journal of Computer-Mediated Communication; Information Sciences)* or political science publications (e.g. *Journal of Public Deliberation, Policy and Politics, ACM Conferences*).

In the remainder of this section, we present the results of our literature review. We discuss the most common design features found in the literature in relation to how each potentially addresses (or not) the challenges of scaling up deliberation online, as identified in Section 2. We include a section on the cross-cutting issues of inclusion and equality, since these are also important challenges in platform design for mass deliberation.

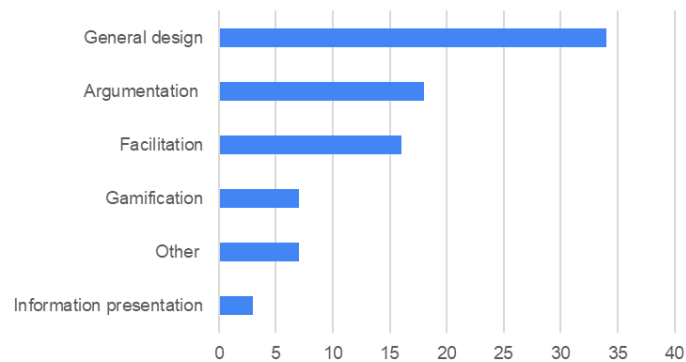

*Figure 2: frequency of design topics*

4.1 Designs dealing with the challenge of large-scale argumentation

The highest number of publications we found were about the design of argumentation tools for deliberation. Argumentation (where rational dialogue is considered to be the basis of conversation), plays a central role in deliberation (Fishkin, 2009). During a deliberation, individuals are expected to use reasoning to produce and evaluate arguments from different perspectives (Mercier and Landemore, 2012). Argumentation tools may speed up the deliberative process by allowing synthesis of large amounts of input when many people are involved in a discussion. These tools help make it easier to understand the arguments of other participants and may therefore promote a more reflective, reasoned discussion online.

Two common formats of argumentation are argument mapping and issue mapping (Kunz and Rittel, 1970). Other models include Bipolar Argument Frameworks (Cayrol and Lagasquie-Schiex, 2005), those using the Argument Interchange Format (Chesnevar et al., 2006), and various other custom models specific to the use case of the article. So far, no single model is accepted by all. Argument mapping makes the process of producing arguments and their interaction explicit (Kirschner et al., 2012). Here, the interplay of claims and premises are mapped into a structured format, for instance into a directed graph. However, no single model of argumentation exists, and different theories exist for the structured format (Van Eemeren et al., 2013; Reed, 2010).

Issue Mapping aims to achieve a similar explicit structure of the content of a debate, but on a more abstract level, in order to create a shared understanding of the problem at hand. Instead of focusing on merely the logical structure of individual utterances, a debate is mapped out in terms of ideas, positions and arguments (Conklin, 2005) using an Issue-Based Information System (IBIS). While not without its own criticism (Isenmann and Reuter, 1997), IBIS remains a highly popular syntax in recent digital deliberation platforms. This popularity is reflected in the studies included in our search: 8 out of the 18 articles that focus on argumentation design employ IBIS in their platforms.

One of the benefits of traditional social media is that anyone can contribute. In large-scale deliberative platforms, this benefit would ideally be preserved when moving to argument-based systems. If an argument maps is easy to browse, enthusiastic reactions of those proficient with the interface can be expected (Liddo and Shum, 2013). However, an argument map restricts social interaction between users, and may introduce a significant learning curve. Iandoli et al. (2016) show that these factors have a negative impact on user's experience of platforms with argument maps in terms of mutual understanding, perceived ease of use and perceived effectiveness of collaboration. Similarly, Gürkan et al. (2010) found that users required significant moderation support to input new ideas that fit the formalization of the argument map, and that user activity, such as "normal" conversations, moved outside of the deliberative platform. To preserve the social interaction, some platforms mix the argumentation map with traditional conversation-based comments (Fujita et al., 2017; Gu et al., 2018; Velikanov, 2017). To make sure everyone can contribute to the discussion, Velikanov (2017) proposes the use of argumentative coaches. In an attempt to provide further insight into the discussion dynamics, metrics surrounding turn-taking, as well as high level thematic information can be added to these overviews (El-Assady et al., 2016). However, these systems are relatively new, and need more evaluation.

Yang et al. (2021) show that training algorithms to merely define what constitutes a positive, neutral and negative argument (to train automated facilitators that act upon that information), may seriously constrain the space and diversity of opinions. [1] Furthermore, an assumption that unrefuted arguments are winning may unduly influence which arguments are of high quality. Even if no refuting reply is present, it does not necessarily mean that the argument is sound (Boschi et al. 2021).

4.2 Designs dealing with facilitating large numbers

The second highest number of publications we found were about the design of facilitation tools, or automated facilitators to support online mass deliberation. Automated techniques (e.g. algorithms involving machine learning and natural language processing) can assist moderators with tedious tasks, improving discussions, saving time and give more equal voice to less willing participants. While complete automation, at this point, is still infeasible and will probably impact the quality of the discussion, platforms have been experimenting with such methods, with varying results (Gu et al., 2018).

Studies dealing with human facilitation in online platforms looked at the impact of human moderation on e.g. conflict (Beck et al., 2018), or perceived fairness and legitimacy (Perrault and Zhang, 2019). Conflict may be managed through interface design that reveals moderators beliefs and values before a discussion (Beck et al., 2018). Excessive levels of moderation, however, may reduce perceived fairness and legitimacy because it can lead to self-censorship or the exclusion of under-represented populations. Perrault and Zhang (2019) suggest using crowdsourced moderation to mitigate such negative effects.

Since human moderators may suffer from their own bias or shortcomings, automated facilitation techniques are an important new avenue of research. Epstein and Leshed (2016) found that online human facilitators' main activities are managing the stream of comments and interacting with comments and commenters. Research on the use of automated facilitation techniques for online deliberation platforms relates to either facilitator assistance tools, which support human facilitators in their roles, or algorithms that completely replace human facilitators.

---

[1] Yang et al. (2021) define: A <u>positive case</u> contains a discussion topic or point that leads to positive discussion and towards consensus. Usually such a post will be followed by several supporting posts. A <u>neutral case</u> represents an ordinary post or opinion that is supportive or objective to the topic opinion in online discussion. It may not play a vital role in leading to forum consensus, but will play a secondary role in the discussion. A <u>negative case</u> represents a post or opinion that contains bad words or detouring information that may distract topic posts or main ideas, even causing flaming or obstruction in online discussion.

Tools to assist facilitation tasks may include making sense of large discussions (Zhang, 2017), discussion summaries, recommending contextually appropriate moderator messages (Lee et al., 2020), visualizing dialogue quality indicators (Murray et al., 2013), structuring an issue-based information system (Ito et al., 2020) or analysing deliberative quality (Fournier-Tombs and Di Marzo Serugendo, 2020). Other tools allow crowd-based idea or argument harvesting (Fujita et al., 2017) or using the machine learning technique, case-based reasoning (CBR), to promote better idea generation, smooth discussion, reduce negative behaviour and flaming and provide consensus-oriented guidance (Yang et al., 2019). In that way, the automated facilitation agent extracts semantic discussion structures, generates facilitation messages, and posts them to the discussion system, while the human facilitator can primarily focus on eliciting consensual decisions from participants (Yang et al., 2021).

Algorithms can also completely replace facilitators and take on the roles of initiating discussions, informing the group, resolving conflict, or playing devil's advocate (Alnemr, 2020). For instance, (Fishkin et al., 2018) created a platform with an automated moderator "bot" which enforces a speaking queue for participants. The platform also incorporates nudges to encourage participants to follow an agenda. As well as this, active speakers are transcribed in real time and monitored for offensive content. Participants can give the bot feedback about whether to block a user or advance the agenda. Artificial discussion agents can also speed up the learning process for participants on complex issues, help participants to better engage with each other, especially those with opposing views. They might also mine new topics for the discussion from the conversations (Stromer-Galley et al., 2012). For example, Wyss and Beste (2017) designed an artificial facilitator (Sophie) for asynchronous discussion based on argumentative reasoning. They added the automatic facilitator to an asynchronous forum (in an argument tree format) with the purpose of creating and accelerating feedback loops according to the argumentative theory of reasoning. They theorised that this would force participants to recognize flaws in their personal reasoning, motivate them to justify their demands, and ensure that they consider the viewpoints of others. Yang et al. (2021) trained their algorithm on solutions human facilitators used in online discussion systems in the past. This means that the algorithm learns how to identify a given 'problem' in a discussion (e.g. "a post or opinion that contains bad words or detouring information that may distract topic posts or main ideas") and then applies a 'solution' (e.g. "remove such a post from discussion or hide it in order to smooth the forum discussion").

While tools and algorithms are shown to be highly useful and reduce the workload of human facilitators in large-scale deliberations, it is important that we weigh up replacing human bias with the inherent bias of the algorithms behind automated moderators. Nonetheless, only a few authors in the literature review take up this discussion. Alnemr (2020) critiques the developments of Stromer-Galley et al. (2012) and Wyss and Beste (2017), noting that algorithms underpinning automated facilitation lack the ability to use discretion as humans do, about how to ensure inclusion, and to enforce certain deliberative norms in a sensitive way. She argues that expert-driven design does not allow citizens to define and agree on deliberative ideals and that instead these ideals are imposed on them by the algorithm, whose design is not transparent.

In the process of facilitating, moderators may also impose their own biases into the conversation. While users themselves could be tasked with inputting their ideas into an argument map (Pingree, 2009), the mapping procedure involves multiple cognitively challenging tasks, increasing the risk for errors. Mappers (those doing the mapping) are expected to keep up with the conversation, while keeping close attention to the map and perform clarification interactions with the discussants. In a review of Finnish moderator tools in various platforms, Ruckenstein and Turunen (2020) warn that some designs may force moderators to operate more like machines and prevent them from using their own skills and judgement, leaving them frustrated in their work. Wyss and Beste (2017) caution that, when designing (automated) facilitators, it is crucial to find a balance between interfering and not interfering in the discussion and this can depend on the context. More research in this regard is clearly needed.

4.3 Designs dealing with lost social cues and replication of social interactions

4.3.1 Gamification

Gamification of deliberation platforms has gained attention in recent years, showing potential to engage more citizens in online deliberation as well as in building empathy and improving civility. Hassan (2017) hopes that gamification would increase low levels of civic participation, and motivate public officials and government agencies to engage in digital deliberative projects as well as reducing negative sentiments towards such projects. She assumes that if more citizens are encouraged to participate in deliberations and perceive their involvement as meaningful, it would not only increase the number of citizens in the governance of their communities, but also help to convince public officials that it can improve governmental decision-making (Hassan, 2017). Beyond engagement and activation, gamification is seen to improve digital deliberations by promoting social connection and empathy, improving self-esteem and social status, argue Gastil and Broghammer (2021). To create empathy for example, the authors suggest that a deliberation bot could ask participants to reflect on the needs and experiences of fictionalized residents of their community or discuss fictional scenarios based on real residents.

The functionalities of gamified environments for deliberation in the literature are many. Some use reward-based approaches to award users based on their activities, or stimulating inter-user competition (Hassan, 2017) others may use collaborative reward structures that provide incentives for reaching a consensus or taking collective action (Gastil and Broghammer, 2021). Others again try to steer users to formulate opinions that fit the discussion phase and to shorten the response time of answering questions (Ito, 2018), or to promote civility among users (Jhaver et al., 2017).

Popular reward functions that have been applied in digital deliberations are deltas [2] (Jhaver et al., 2017), badges [3] (Bista et al., 2014), award scores (Yang et al., 2021) or discussion points/virtual money (Ito, 2018). Gastil and Broghammer (2021) provide a further comprehensive list of gamification mechanisms suitable for online tools that include deliberation, such as competition over resources; material rewards; scarcity of rewards; artificial time constraints; chats; peer rating system, shared narratives; learning modules; progress bars; leaderboards and missions.

Few case studies have been conducted so far that measure the effects of gamification. Bista et al. (2014) for example used a set of badges as awards for a gamified community for Australian welfare recipients who were encouraged to communicate with each other and with the government. According to the authors, gamification increased participant's reading and commenting actions, and helped the researchers to obtain accelerated feedback on the 'mood' in the community. A recent review found however that it is difficult to conclude the impact of gamification in online deliberations, since gamification is never equally experienced and appreciated by all sorts of social groups in the same way (Hassan and Hamari, 2020).

Jhaver et al. (2017), who studied the subreddit ChangeMyView, observed that gamification mechanisms in the form of deltas increase civility and politeness between users. This is particularly obvious for new members of the community as well as for high-scorers. Even though there is a competitive element that allows high performers to compete with one another, the focus of the community is on meaningful conversations, and thus, according to the authors, the users are not explicitly judged by their delta scores. These findings are in line with Hamari et al. (2014) that conclude that positive outcomes from gamification depend on the context and the characteristics of users.

Downsides of gamification methods have also been reported: features that allow participants to build reputation via upvoting for instance. While having voting mechanisms may seem democratic, in reality this could be far from the case. Humans tend to be highly influenced by the opinions of others: for instance, studies show that comments with an early upvote are 25% more highly rated in the end, regardless of their quality (Maia and Rezende, 2016; Muchnik et al. 2013). Some users might also be driven solely to achieve high scores and gaming the gamification rather than to actually contribute to the deliberation goal (Bista et al., 2014). Such occurrences might lead designers to 'hide' the full rules of calculating points to participants, which then raises issues of transparency.

Gamification can be an important addition to developing digital deliberation interfaces, and if embedded with social norms, and moderation mechanisms, it can improve a digital space to foster productive discourse. Competition generated by gamification should never be upfront. Nor should it be unidirectionally focused on those members who are already achieving strong results (Jhaver et al., 2017). Civic gamification designers should focus on understanding the psychology (intrinsic motivation) of why citizens would participate in an online deliberation in the first place, for gamification efforts to provide value (Hassan, 2017). Moreover, gamification leads to more data processing, which does not necessarily mean that this data informs decision-makers in ways that represent public opinion. Cherry-picking and biases might occur in terms of privileging what they see as 'facts', and 'evidence' such as 'user engagement', 'likes' or 'scores' etc., because decision-makers succumb commonly more to these types of aggregated data than to qualitative attributes (Johnson et al., 2017).

---

[2] Deltas are bonus tokens that are awarded if someone had a change of view or mentioned a change of view in their response.

[3] Badges are digital emblems for altruistic and honest engagement actions.

4.3.2    Anonymity and identifiability

Choosing between identification or anonymity in digital deliberations creates a number of trade-offs. Anonymity can imply a loss of accountability and respectfulness or may negatively influence civility in online discussions (Sarmento and Mendonça, 2014). Even if anonymity decreases barriers to entry, it can lead to less reflection and more extreme contributions. Reducing anonymity may have a positive effect on respectfulness and thoughtfulness (Coleman and Moss, 2012) and increases transparency, but has a possibly negative effect on engagement (Rhee and Kim, 2009) — people tend to contribute less to the discussion overall when they are identifiable. This can also raise questions of privacy on the platforms. Gonçalves et al. (2020) indicate how easy it is to build a profile of each participant by following, or studying the tracks they unconsciously leave behind while discussing, proposing ideas or interacting with others as well as their behaviour in a gamified environment.

In addition, it has also been found that when users share a common social identity in an online community, they are more susceptible to group influence and stereotyping, despite participant anonymity (Jhaver et al., 2017). Incorporating the focus theory of normative conduct (Cialdini et al., 1990) into platform cue design may improve online dialogue by enforcing three different types of norm: descriptive, injunctive and personal. Respectively, these norms motivate behaviour by promising either rewards or sanctions externally imposed by others; by providing frequency information about the behaviour of others and reflecting individuals' commitment with their internalized values. Manosevitch et al. (2014) show that visual/cognitive cues like banners can prime participants to be aware that they are in a deliberative context. Such cues may encourage e.g. reflection, considering a range of opinions, and being true to one's self and hence improve the deliberative quality of the discussion.

4.4 Designs dealing with inclusion and equality concerns

To begin with, few studies deal with the issue of gender, or gendered behaviour in relation to facilitation approaches or the design of automated facilitation tools in deliberative software. One study, (Kennedy et al., 2020) found that the gender of human facilitators impacted discussion outcomes, that a text-based environment may be favourable to female participants since it prevents interruptions, but that non-white and participants over 65 were less active in such discussions.

Three publications deal with the topic of learning styles or argumentative capability of participants: Velikanov (2017) makes some preliminary technical and non-technical design suggestions (e.g. argument coaches or facilitator incentives) to address the topic of differing levels of factual preparedness and argumentative capability as well as the issue of linguistic differences. Epstein and Leshed (2016) described heuristics that can be used by moderators to respond to different levels of participatory literacy and hence improve the overall quality of the comments. Brinker et al. (2015) found that using mixed media providing favourable outcomes for developing social ties, building mutual understanding and encouraging reflection on values as well as facts, and formulating arguments. In their study, Wyss and Beste (2017) show that the effectiveness of the automated facilitator in creating favourable learning conditions for participants depends also on their personality traits, e.g. on whether participants were conflict-avoidant or had a high need to evaluate.

Several studies examine the impact of social cues on social dynamics. With anonymity, a more egalitarian environment is possible since people feel more freedom to express their honest, even if unpopular, point of view. Removing visual cues of gender, age and race promotes equal treatment of individuals (Kennedy et al., 2020). Harmful social dynamics are reduced and people stay more focused on the task at hand (Iandoli et al., 2014). Anonymity can also allow civil servants or people with neutrality obligations to participate. Asynchronous discussion is also a way to "level the playing field" between the more and less informed public (Neblo et al., 2010), and has been shown to encourage women to participate more by removing interruptions (Kennedy et al., 2020). While conducting field experiments for gamification in online deliberation, Johnson et al. (2017) found issues with equality, turn taking, providing evidence in discussions, elaboration of stakeholder opinions and documentation. After inserting a turn-taking mechanism, combined with restricting the topic of conversation through "cards" prevented talks from going off-topic. Moreover, reasoning prompts improved empathy and respect, as well as the change of viewpoints following some discussions. This also reduced the workload for facilitators, as groups of participants regulated themselves according to the rules of the game.

With regard to the design process itself, it is especially interesting, that of the 62 design case studies we analysed (see Appendix A), around half had limited or no descriptions of the users or participants who tested or collaborated in the software design. The remaining case studies provided some information on one or more of the following: number of participants or users of the software, (mean) age, gender balance, ethnicity, education levels, political affiliation and profession of participants.

Just 33 case-studies mentioned the number of participants or test users, with numbers ranging from small groups to several thousand users. Only 13 studies provided details about the gender balance of participants. In 9 of these studies, females made up less than 50% of participants. Only 3 studies reported information about the race or ethnicity of participants. The education levels of participants, when reported, was nearly always university level, probably because in the majority of cases where this information was reported, the test users were university students or staff. Only 10 studies reported the age range of participants. While a few studies used samples representative of the general population, for the most part, the age range for participants in these studies skewed young (under 40). In brief, where reported, the majority of test user groups seem to be young, male and WEIRD (Western, Educated, Industrialized, Rich, Democratic).

## 5   Discussion and future research avenues

Six years ago, in Friess and Eilders' online deliberation review (2015), the number of mentions about automated facilitation, argumentation mining, algorithms, or gamification was exactly zero. Much has changed since then. As this literature review shows, the development and experimental use of mass deliberative platforms has leapt forward in recent years, and has not only seen increasing scholarly attention, but also a rise of tech-start-ups. We have found references to 106 different deliberative online platforms in our review (see Appendix B).

As we have focused on designers and their progress in addressing the challenges to scaling up deliberation, our review highlights that automated facilitation and argumentation tools receive by far the most attention in the literature surveyed. Moreover, we recognize, this is a move towards reliance on technical methods due to the large number of users and information produced and a possible loss of quality in deliberation. Since argumentation tends to be regarded as a key goal of deliberation, argumentation mining and mapping tools are currently the focus of extensive research and experimentation. An interesting discrepancy of this process is that while years of deliberation research has gone to great pains to conceptualize what makes a good argument, or what is crucial for deliberative quality or how deliberation can be applied in a given context or culture, todays tools do not necessarily reflect this. Current efforts that train an algorithm to detect and label and argument often resort to pre-existing popular syntax, such as Case-Based Reasoning (CBR), Issue-Based Information System (IBIS), Bipolar Argument Framework (BAF) or Argument Interchange Format (AIF). While we have not compared the syntax behind crowd deliberative platforms to deliberative ideals in this research in detail, it would be an avenue of research to investigate if these syntaxes form a new understanding of deliberative ideals and if they are at cross-roads with older definitions. Moreover, a perspective found to be missing in our literature search relates to a critique of logical argument structuring (e.g. Durnova et al. (2016), which argues that a discursive approach may be more appropriate since it recognises the subjectivity of actors, their different forms of knowledge and interpretations from which they create meaning. It is shown, for example, that rather than sharing facts, people are often more convinced by stories about personal experiences (Kubin et al., 2021), which suggests that deliberative software that relies on the exchange of arguments, or factual statements may not achieve the other desired goals of deliberation such as building respect for different points of view or empathy.

Furthermore, as our review shows, much work remains to be done in relation to tools that assist with cognitively challenging tasks involved in mapping the arguments, deciding which arguments should win, teaching diverse participants to use argumentation itself and allowing for social interaction in parallel, so that such platforms can be as attractive as popular social media. Too heavy a focus on designs that emphasise argumentation may exclude people with limited argumentation and rhetorical skills or that prefer other types of expression. Populations with special needs may also be excluded, including older adults, children, people with disabilities or who are illiterate (López and Farzan, 2017). Language may also be an issue in that if not all participants speak the same first language, this may lead to further inequalities (Velikanov, 2017). Our analysis of the literature suggests that most research on deliberative platforms uses highly educated test users, likely from a comfortable economic background. However, in practice, difficulties may also arise with including certain social groups (minorities, the poor or less-educated) in deliberations (Asenbaum, 2016). Use of the internet is influenced by less than obvious factors such as social class or sense of self-efficacy, which often affect people of colour or women (López and Farzan, 2017) or motivation, access to equipment, materials, skills or the social, political and economic context (Epstein et al., 2014). Some citizens may believe complex issues are beyond their expertise and hence defer decisions to be made by experts (Font et al., 2015) and people may need training or education to help develop the capabilities to participate in a deliberative setting (Beauvais, 2018). The skills of reformulating provocative content, mirroring perspectives, posing circular questions or playing the devil's advocate, are only a few of the important techniques a trained facilitator can employ to support argumentatively disadvantaged participants or to redirect a discussion that reaches a dead end. In our literature review, we find that automatic facilitation tools currently lack the functionality for such nuanced interventions on a large scale.

Culture also influences the appropriateness of using argumentation. In particular, Western cultural and methodological standards currently dominate deliberation research (Min, 2014). However, certain cultures may favour different deliberation or argumentation styles e.g. cultures may be consensus-based or adversarial (Bächtiger and Hangartner, 2010). For example, Confucian societies may value social harmony over public disagreement. Other studies (Becker et al., 2019; Shi et al., 2019) have highlighted the positive impact of partisans and polarized crowds. Under the right design principles, these groups produce high-quality outputs. Moreover, in Muslim countries men and women may deliberate separately (Min, 2014). An in societies where ethnic, religious, or ideological groups have historically each found their own identity in rejecting the identity of the other, deliberation needs to find common ground first (Dryzek et al., 2019). Our analysis of the geographical distribution of studies shows that a high proportion of research on the design of deliberative platforms takes place in Western, developed countries, and rarely reports the ethnicity of the participants, which will doubtless have implications for the design of future platforms and should be a concern.

Automated facilitation tools also receive a lot of attention and are being used of to tackle problems related to scaling up deliberation such as managing comments, maintaining respectfulness, encouraging participation, making sense of vast amounts of information or monitoring the discussion progress and quality. However, while such tools may aim to circumvent human facilitator bias, issues of algorithmic bias are hardly discussed in the publications we found. There is very little discussion about the values or interpretation of deliberative democracy that underpins automated facilitation tools.

Little attention is paid in the literature surveyed to the effects of gendered behaviour (Afsahi, 2020), social inequality (Beauvais, 2018) or the implications of different communication or learning styles (Siu, 2017), all of which can influence inclusiveness and equality in deliberation. Communication style and gendered behaviours are carried over into deliberative settings. The communication skills and style of expression required in deliberative settings tends to be characteristic of higher income white males and may not be characteristic of all social groups, leaving them at a disadvantage. For example, high-resource and digitally engaged individuals are generally more capable and active in discussions (Himmelroos et al., 2017; Sandover et al., 2020), leaving less-privileged, less-educated, or perhaps illiterate participants at a disadvantage in discussions with the more privileged, better educated, and well-spoken (Hendriks, 2016; Siu, 2017). Furthermore, men are more likely to interrupt or ignore women and speak for longer and more often than women, who tend to be less assertive and more accommodating in deliberations (Siu, 2017). Women's contributions may be marginalized or sexualised due to gender hostility (Kennedy et al., 2020). Women also tend to be more conflict avoidant and less willing to engage in argumentation required for deliberation, hence they may require a particular style of facilitation to ensure their inclusion (Afsahi, 2020). Our analysis of the literature suggests that women tend to be underrepresented in the groups of users that test deliberative platforms during the design process. This is a concern and should be considered in future design studies.

Although facilitation and argumentation tools may overcome certain challenges to mass deliberation, other design formats are also important. Substantial attention is now given to gamification features in the literature, which address a different issue, that of encouraging participation online and fostering feelings of connection and empathy. Some research has begun on how to accommodate different communication and learning styles, the use of gamification as an addition to argumentation or a way to build rapport between online participants. Experimental work on different gamification designs (e.g. competitive design, collaborative design, etc.) shows promise, in particular in increasing engagement. There are further indications that gamification can increase engaged interaction when it is designed around contextual and user-specific means, rather than around what is popular in gamification research (Hassan and Hamari, 2020). Further research on how gamification can promote different communication styles of different deliberative ideals is needed. Until now, literature fails to demonstrate a wide array of comparative studies (Hassan, 2017) and it is unclear how different user-groups utilize gamification functions to their advantage in certain contexts.

In order to ensure that a fair interpretation of deliberative ideals is reflected in platform design, rather than an expert driven interpretation (Alnemr, 2020), the use of design methodologies such as participative (PD) or value-sensitive design (VSD) is recommended. Design methodologies that involve stakeholder participation are more likely to reflect the needs and values of the intended users. However, the selection of stakeholders must be carefully executed. Universal design (see López and Farzan, 2017) may be particularly interesting, since it involves the design of products and environments to be usable by all people, to the greatest extent possible, without the need for adaptation or specialized design.

Similar to small-scale online deliberation, as pointed out by Friess and Eilders (2015), research about outcome factors, such as opinion change, trust (on the individual level) or quality and legitimacy (on the collective level) are also still largely missing in crowd size deliberations. However, the amount of generated data by mass deliberative platforms is already estimated to be huge. It is thus likely that in a short period, outcome measurements will be very likely to follow. Caution has to be exercised, however, when participants' behaviour is surveyed, paternalised or nudged into a certain direction, to achieve what Hassan and Hamari (2020) call, the "creation of good citizens".

# 6 Conclusion

This paper reviewed the literature on design features of online deliberation platforms with regard to how they address the challenges of scaling up deliberation to crowd size. We found that the most commonly studied design features are argumentation tools and facilitation tools. While these tools address certain issues such as the large volume of information and need for structuring arguments, they may neglect the more nuanced requirements of high quality deliberation and possibly reduce inclusion or uptake. Our analysis of the literature shows that the characteristics (age, gender, education, ethnicity/race, etc.) of test users is rarely reported in case studies we found and the design of deliberation tools mainly takes place in a Western context. Based on our findings, deliberative platforms are more likely to reflect the values and needs of a small, unrepresentative, segment of the world's population. The resulting mismatch with the technologies and social groups that use them risks impeding a wider uptake. Some designs that feature gamification or allow anonymity or asynchronous participation may address certain issues like building empathy or avoiding some types of discrimination in online platforms. However, in general, much research is still needed on how to facilitate and accommodate different genders or cultures in deliberation, how to deal with the implications of pre-existing social inequalities, how to deal with differences in cognitive abilities and cultural or linguistic differences or how to build motivation and self-efficacy in certain groups. Many studies do not yet bridge the required disciplines. Even if the topics of some publications could be considered as traversing disciplines, they may not be reaching a broad enough audience. Our findings highlight the importance of breaking the silos between disciplines, but in particular to better integrate gender studies, psychology, anthropology, psychology or the discipline of design itself in order to create attractive and inclusive deliberation platforms going forward. It will also be crucial to examine the impact of new design features on the uptake of the software tools as well as the quality of deliberation.

## Funding

This work has been supported by the Dutch research council and (partially) funded by the Hybrid Intelligence Center, a 10-year programme funded by the Dutch Ministry of Education, Culture and Science through the Dutch Research Council.

Gastil, J. and Broghammer, M. (2021). Linking theories of motivation, game mechanics, and public deliberation to design an online system for participatory budgeting. *Political Studies*, 69(1):7–25.

Gastil, J. and Meinrath, S. D. (2018). Bringing citizens and policymakers together online: Imagining the possibilities and taking stock of privacy and transparency hazards. *Computer*, 51(6):30–40.

Gold, V., USS, B. P., El-Assady, M., Sperrle, F., Budzynska, K., Hautli-Janisz, A., and Reed, C. (2018). Towards deliberation analytics: stream processing of argument data for deliberative communication. In *Proceedings of COMMA Workshop on Argumentation and Society*, pages 1–3.

Gonçalves, F. M., Prado, A. B., and Baranauskas, M. C. C. (2020). Opendesign: Analyzing deliberation and rationale in an exploratory case study. In *ICEIS (2)*, pages 511–522.

Goodin, R. E. (2000). Democratic deliberation within. *Philosophy & Public Affairs*, 29(1):81–109.

Gordon, E., Haas, J., and Michelson, B. (2017). Civic creativity: Role-playing games in deliberative process. *International Journal of Communication*, 11:19.

Gordon, E. and Manosevitch, E. (2011). Augmented deliberation: Merging physical and virtual interaction to engage communities in urban planning. *New Media & Society*, 13(1):75–95.

Gordon, E., Michelson, B., and Haas, J. (2016). @ stake: A game to facilitate the process of deliberative democracy. In *Proceedings of the 19th ACM Conference on Computer Supported Cooperative Work and Social Computing Companion*, pages 269–272.

Grönlund, K., Himmelroos, S., et al. (2009). The challenge of deliberative democracy online–a comparison of face-to-face and virtual experiments in citizen deliberation. *Information Polity*, 14(3):187–201.

Gu, W., Moustafa, A., Ito, T., Zhang, M., and Yang, C. (2018). A case-based reasoning approach for automated facilitation in online discussion systems. In *2018 Thirteenth International Conference on Knowledge, Information and Creativity Support Systems (KICSS)*, pages 1–5. IEEE.

Gürkan, A. and Iandoli, L. (2009). Common ground building in an argumentation-based online collaborative environment. In *Proceedings of the International Conference on Management of Emergent Digital EcoSystems*, pages 320–324.

Gürkan, A., Iandoli, L., Klein, M., and Zollo, G. (2010). Mediating debate through on-line large-scale argumentation: Evidence from the field. *Information Sciences*, 180(19):3686–3702.

Gutmann, A., Thompson, D. F., et al. (2004). *Why Deliberative Democracy?* Princeton University Press.

Hamari, J., Koivisto, J., and Sarsa, H. (2014). Does gamification work?–a literature review of empirical studies on gamification. In *2014 47th Hawaii international conference on system sciences*, pages 3025–3034. Ieee.

Hamlett, P. W. and Cobb, M. D. (2006). Potential solutions to public deliberation problems: Structured deliberations and polarization cascades. *Policy Studies Journal*, 34(4):629–648.

Hartz-Karp, J. and Sullivan, B. (2014). The unfulfilled promise of online deliberation. *Journal of Public Deliberation*, 10(1):1–5.

Hassan, L. (2017). Governments should play games: Towards a framework for the gamification of civic engagement platforms. *Simulation & Gaming*, 48(2):249–267.

Hassan, L. and Hamari, J. (2020). Gameful civic engagement: A review of the literature on gamification of eparticipation. *Government Information Quarterly*, 37(3):101461.

Hendriks, C. M. (2016). Coupling citizens and elites in deliberative systems: The role of institutional design. *European Journal of Political Research*, 55(1), 43–60.

Himmelroos, S., Rapeli, L., & Grönlund, K. (2017). Talking with like-minded people—Equality and efficacy in enclave deliberation. *Social Science Journal*, 54(2), 148–158.

Iandoli, L., Quinto, I., De Liddo, A., and Buckingham Shum, S. (2016). On online collaboration and construction of shared knowledge: Assessing mediation capability in computer supported argument visualization tools. *Journal of the Association for Information Science and Technology*, 67(5):1052–1067.

Iandoli, L., Quinto, I., De Liddo, A., and Shum, S. B. (2014). Socially augmented argumentation tools: Rationale, design and evaluation of a debate dashboard. *International Journal of Human-Computer Studies*, 72(3):298–319.
16

Appendix A: Reviewed case-studies

| Paper | Topic | Country | User description |
|---|---|---|---|
| Aragón, P., Gómez, V., & Kaltenbrunner, A. (2017). Detecting Platform Effects in Online Discussions. Policy and Internet, 9(4), 420–443. | Argumentation | Spain | n/a |
| Beck, J., Neupane, B., & Carroll, J. M. (2018). SocArXiv Managing Conflict in Online Debate Communities : Foregrounding Moderators ' Beliefs and Values on, (1). | Human facilitation | USA | n/a |
| Bista, S. K., Nepal, S., Paris, C., & Colineau, N. (2014). Gamification for online communities: A case study for delivering government services. international Journal of Cooperative information Systems, 23(02), 1441002. | Gamification | Australia | n/a |
| Boschi, G., Young, A. P., Joglekar, S., Cammarota, C., & Sastry, N. (2019). Having the Last Word: Understanding How to Sample Discussions Online, 1–27. | Argumentation | UK | n/a |
| Brinker, D. L., Gastil, J., & Richards, R. C. (2015). Inspiring and Informing Citizens Online: A Media Richness Analysis of Varied Civic Education Modalities. Journal of Computer-Mediated Communication, 20(5), 504–519. | Information presentation / media types | USA | gender, race, age, political leaning |
| Davies, T., & Chandler, R. (2012). Online deliberation design: Choices, criteria, and evidence. In T. Nabatchi, J. Gastil, M. Leighninger, & G. M. Weiksner (Eds.), Democracy in Motion: Evaluating the Practice and Impact of Deliberative Civic Engagement (pp. 103–134). Oxford Scholarship Online. | General design | USA | n/a |
| Davies, T., O'Connor, B., Cochran, A. A., Effrat, J. J., Parker, A., Newman, B., & Tam, A. (2009). An Online Environment for Democratic Deliberation: Motivations, Principles, and Design. Online Deliberation: Design, Research, and Practice, 5(1), 275–292. Retrieved from http://www.stanford.edu/~davies/deme-principles.pdf | General design | USA | profession |
| De Cindio, F., & Stortone, S. (2013). Experimenting liquidfeedback for online deliberation in civic contexts. Lecture Notes in Computer Science (Including Subseries Lecture Notes in Artificial Intelligence and Lecture Notes in Bioinformatics), 8075 LNCS, 147–158. | Argumentation | Italy | age, gender |
| De Liddo, Anna and Buckingham Shum, Simon (2010). Capturing and representing deliberation in participatory planning practices. In: Fourth International Conference on Online Deliberation (OD2010), 30 Jun - 2 Jul 2010, Leeds, UK | Argumentation | UK | n/a |
| Delborne, J. A., Anderson, A. A., Kleinman, D. L., Colin, M., & Powell, M. (2011). Virtual deliberation? prospects and challenges for integrating the internet in consensus conferences. Public Understanding of Science, 20(3), 367–384. | General design | USA | n/a |
| Dunne, K. (2011). Can online forums be designed to empower local communities? TripleC, 9(2), 154–174. | General design | UK | n/a |
| Easterday, M. W., Kanarek, J. S., & Harrel, M. (2005). Design Requirements of Argument Mapping Software for Teaching Deliberation. Education, (January 2009), 317–324. | Argumentation | USA | n/a |
| Epstein, D., & Leshed, G. (2016). The magic sauce: Practices of facilitation in online policy deliberation. Journal of Public Deliberation, 12(1), 4. | Human facilitation | USA | gender, profession, education |
| Epstein, D., Newhart, M., & Vernon, R. (2014). Not by technology alone: The "analog" aspects of online public engagement in policymaking. Government Information Quarterly, 31(2), 337–344. | General design | USA | n/a |
| Ercan, S. A., Hendriks, C. M., and Dryzek J. S. (2019). "Public deliberation in an era of communicative plenty." Policy & politics 47.1: 19-36. | Reflection spaces | Australia | n/a |

## Appendix B: Mass deliberative platforms

Out of a total of 106 unique platforms, only 11 were mentioned > 1. These are shown below. The functions are Gamification (G), Argumentation (A) and Facilitation (F). A full list is available on request.

| Platform | Count | Function | Academic/Commercial | Active |
|---|---|---|---|---|
| Deliberatorium | 6 | G,A,F | Academic | Yes, but inaccessible |
| COLLAGREE | 5 | A,F | Academic | Yes, merged into D-Agree |
| Regulation Room | 4 | F | Academic | No |
| ConsiderIt | 3 | A | Commercial | Yes |
| Cohere | 2 | A | Academic | Yes, but discontinued development |
| Kialo | 2 | A, F | Commercial | Yes |
| liquidfeedback | 2 | A | Commercial | Yes |
| MOOD | 2 | - | Academic | No |
| PICOLA | 2 | F | Academic | No |
| Reflect! | 2 | - | Academic | Yes |
| @Stake | 2 | G | Academic | Yes |